\pdfoutput=1

\documentclass[11pt]{article}

\usepackage[preprint]{acl}

\usepackage{times}
\usepackage{latexsym}
\usepackage[T1]{fontenc}

\usepackage[utf8]{inputenc}

\usepackage{microtype}

\usepackage{inconsolata}

\usepackage{times}  
\usepackage{helvet}  
\usepackage{courier}  
\usepackage{graphicx} 
\usepackage{natbib}  
\usepackage{caption} 
\usepackage{algorithm}
\usepackage{algpseudocode}
\usepackage{subcaption} 
\usepackage{tabularx} 
\usepackage{amsmath}
\usepackage{amsthm}
\usepackage{amsfonts}
\usepackage{colortbl}
\usepackage{listings}
\usepackage{amssymb}
\usepackage{pifont}
\usepackage{geometry}

\floatstyle{ruled}
\newfloat{listing}{tb}{lst}{}
\floatname{listing}{Listing}
\usepackage[utf8]{inputenc}
\usepackage[T1]{fontenc}
\usepackage{lipsum} 

\usepackage{fontawesome} 
\usepackage[most]{tcolorbox}
\usepackage{listings}
\usepackage{microtype}  

\definecolor{boxbg}{RGB}{248, 250, 252}
\definecolor{boxframe}{RGB}{41, 128, 185}
\definecolor{titlebg}{RGB}{52, 152, 219}
\definecolor{codebg}{RGB}{250, 250, 250}
\definecolor{keyword}{RGB}{41, 128, 185}
\definecolor{string}{RGB}{230, 126, 34}
\definecolor{comment}{RGB}{149, 165, 166}
\definecolor{selfcolor}{RGB}{162, 29, 162} 

\tcbset{
  mypromptbox/.style={
    enhanced,
        breakable,        
    colback=boxbg,
    colframe=boxframe,
    arc=3mm,
    boxrule=0.8pt,
    leftrule=3pt,
    toprule=0.8pt,
    bottomrule=0.8pt,
    rightrule=0.8pt,
    left=8pt,
    right=8pt,
    top=8pt,
    bottom=8pt,
    fonttitle=\sffamily\bfseries,
    coltitle=white,
    colbacktitle=titlebg,
    attach boxed title to top left={yshift=-2mm, xshift=5mm},
    boxed title style={
      sharp corners, 
      rounded corners=southwest, 
      arc=1mm, 
      boxrule=0.8pt
    },
    shadow={2mm}{-1mm}{0mm}{black!40},
    overlay={
      \begin{tcbclipframe}
        \draw[boxframe!30, line width=0.4pt, dash pattern=on 2pt off 2pt] 
          ([xshift=6pt, yshift=-6pt]frame.north west) -- 
          ([xshift=-6pt, yshift=-6pt]frame.north east);
      \end{tcbclipframe}
    }
  }
}

\lstdefinestyle{promptStyle}{
    language=Python,
    basicstyle=\ttfamily\footnotesize\color{black!80},
    keywordstyle=\color{keyword}\bfseries,
    commentstyle=\color{comment}\itshape,
    stringstyle=\color{string},
    backgroundcolor=\color{codebg},
    breaklines=true,
    breakatwhitespace=true,
    frame=single,
    framesep=6pt,
    numbers=left,
    numberstyle=\tiny\color{black!60},
    numbersep=10pt,
    tabsize=4,
    showstringspaces=false,
    captionpos=b,
    xleftmargin=18pt,
    framexleftmargin=18pt,
    framextopmargin=6pt,
    framexbottommargin=6pt,
    rulecolor=\color{boxframe!20},
    postbreak=\raisebox{0ex}[0ex][0ex]{\ensuremath{\color{boxframe}\hookrightarrow\space}},
    aboveskip=12pt,
    belowskip=12pt,
    emphstyle={\color{purple}\bfseries}, 
    emph={import,from,class,def,for,while,if,is,else,elif,return,try,except,
          finally,break,continue,with,as,in,not,or,and,True,False,None},
    morekeywords=[1]{self}, 
    keywordstyle=[1]{\color{selfcolor}\bfseries},
    escapeinside={\%*}{*\%},
    extendedchars=true,
    upquote=true,
    columns=flexible
}

\usepackage{xspace}
\usepackage{multirow}
\usepackage{mathrsfs} 
\newcommand{\ORMind}{{\textsf{ORMind}}\xspace}
\newcommand{\xmark}{\ding{55}}

\title{ORMind: A Cognitive-Inspired End-to-End Reasoning Framework for Operations Research}

\author{
  \textbf{Zhiyuan Wang\textsuperscript{1†}\thanks{Work done as an intern at AI Lab of Lenovo Research. }}, 
  \textbf{Bokui Chen\textsuperscript{1,5}\thanks{Equal contributions.}}, 
  \textbf{Yinya Huang\textsuperscript{3}}, 
    \textbf{Qingxing Cao\textsuperscript{4}\thanks{Corresponding authors.}}, 
  \\ 
\textbf{Ming He}\textsuperscript{2‡},
  \textbf{Jianping Fan\textsuperscript{2}}, 
  \textbf{Xiaodan Liang\textsuperscript{4,5}} \\
 \textsuperscript{1}Tsinghua Shenzhen International Graduate School, Tsinghua University,
 \\
 \textsuperscript{2}AI Lab of Lenovo Research,
 \textsuperscript{3}ETH Zurich,
 \textsuperscript{4}Sun Yat-sen University,
 \\
 \textsuperscript{5}Peng Cheng Laboratory
  \\
  \texttt{{\{wang-zy22, chenbk\}}@tsinghua.edu.cn},
  \texttt{yinya.huang@hotmail.com} \\
  \texttt{heming01@foxmail.com} ,\texttt{caoqx@mail2.sysu.edu.cn}, \\ \texttt{jfan1@lenovo.com}, \texttt{xdliang328@gmail.com}
}

\begin{document}
\maketitle
\begin{abstract}
Operations research (OR) is widely deployed to solve critical decision-making problems with complex objectives and constraints, impacting manufacturing, logistics, finance, and healthcare outcomes. While Large Language Models (LLMs) have shown promising results in various domains, their practical application in industry-relevant operations research (OR) problems presents significant challenges and opportunities. Preliminary industrial applications of LLMs for operations research face two critical deployment challenges: 1) Self-correction focuses on code syntax rather than mathematical accuracy, causing costly errors; 2) Complex expert selection creates unpredictable workflows that reduce transparency and increase maintenance costs, making them impractical for time-sensitive business applications. To address these business limitations, we introduce ORMind, a cognitive-inspired framework that enhances optimization through counterfactual reasoning. Our approach emulates human cognition—implementing an end-to-end workflow that systematically transforms requirements into mathematical models and executable solver code. It is currently being tested internally in Lenovo's AI Assistant, with plans to enhance optimization capabilities for both business and consumer customers. Experiments demonstrate that ORMind outperforms existing methods, achieving a 9.5\% improvement on the NL4Opt dataset and a 14.6\% improvement on the ComplexOR dataset. Code is currently available at \href{https://github.com/XiaoAI1989/ORMind}{https://github.com/XiaoAI1989/ORMind}.

\end{abstract}

\section{Introduction}

\begin{figure*}[htb]
    \centering
    \begin{subfigure}[b]{0.545\textwidth}
        \includegraphics[width=\textwidth,keepaspectratio]{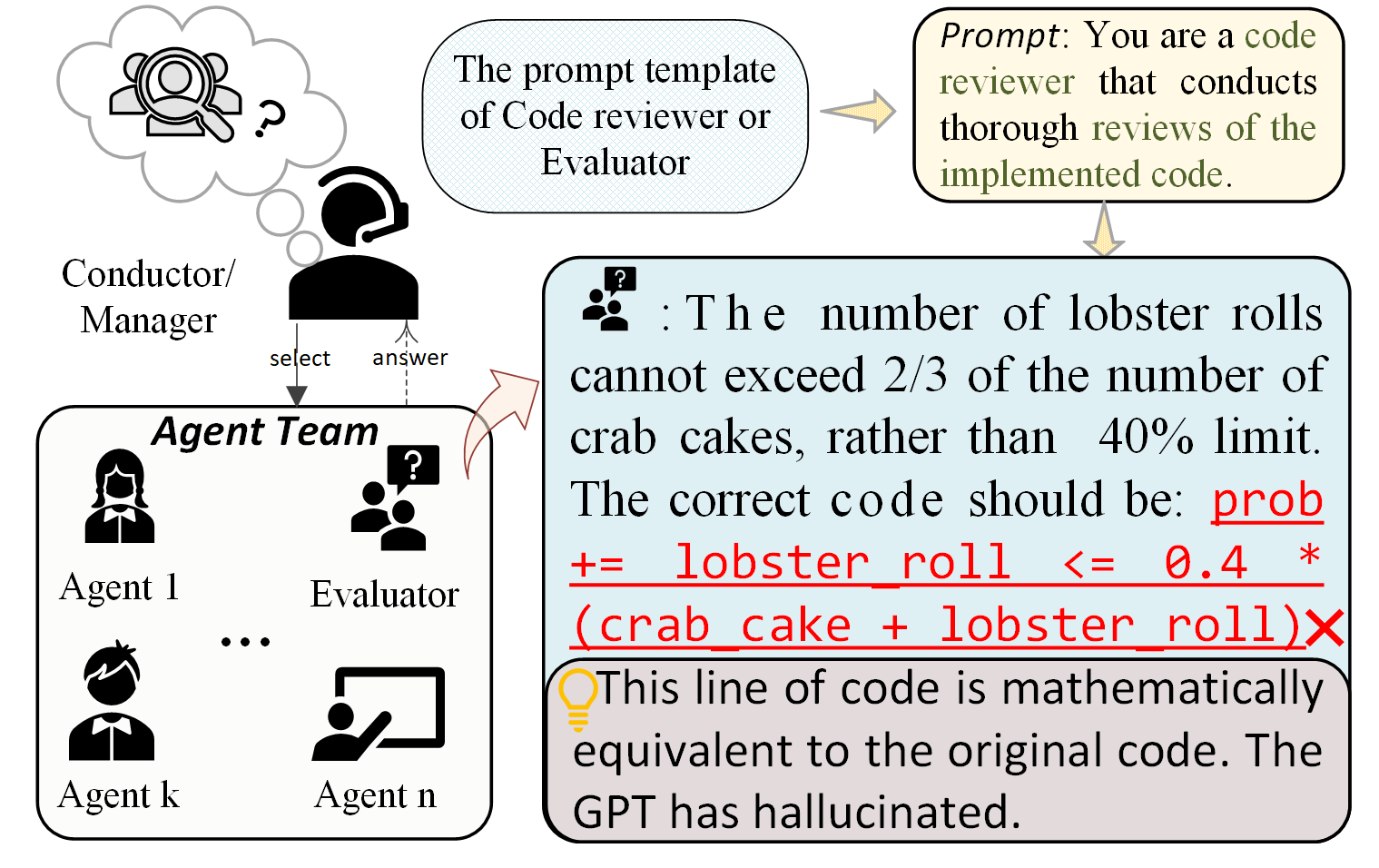}
        \caption{}
        \label{fig:previous_method}
    \end{subfigure}
    \hfill
    \begin{subfigure}[b]{0.435\textwidth}
        \includegraphics[width=\textwidth,keepaspectratio]{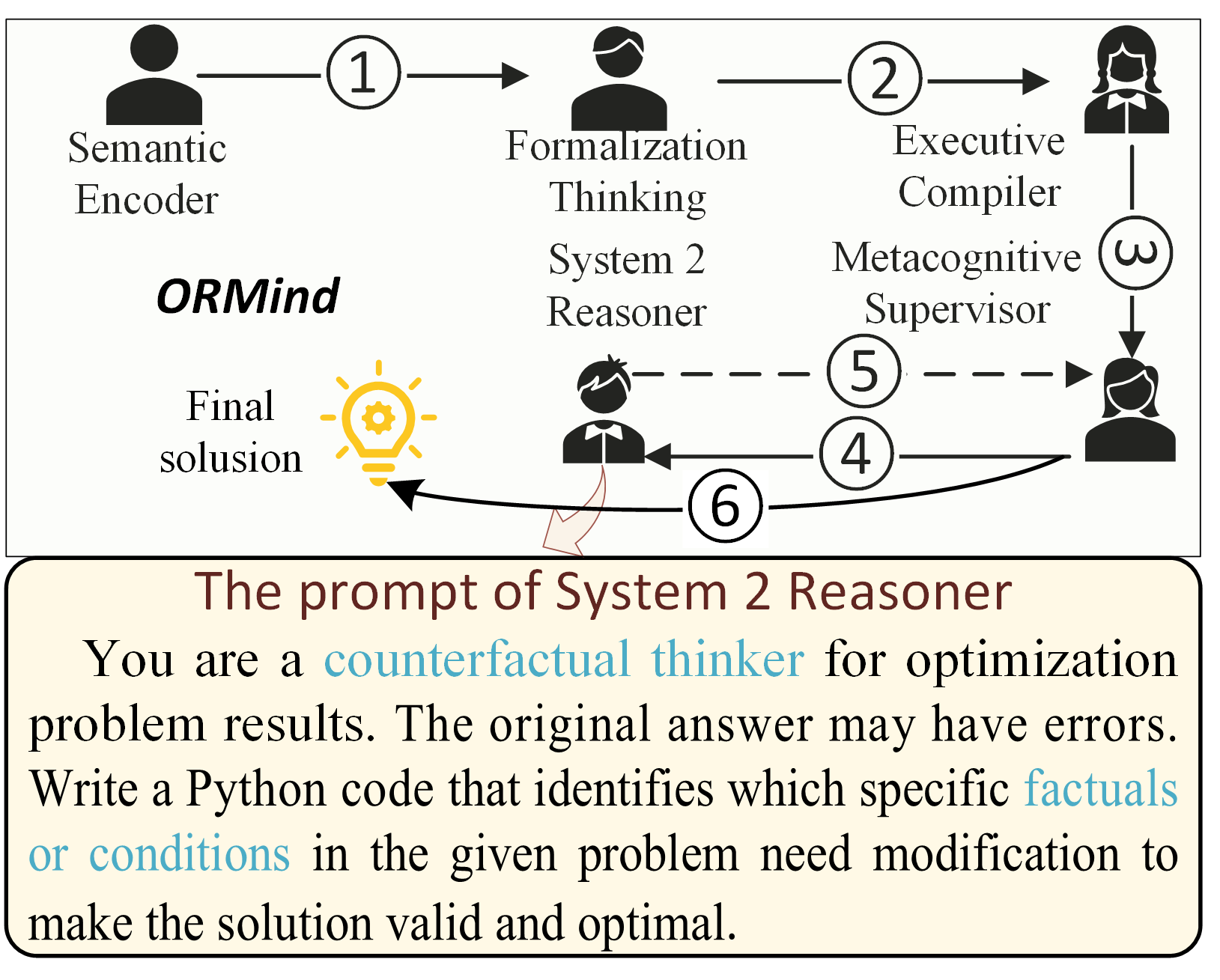}
        \caption{}
        \label{fig:our_method}
    \end{subfigure}

    \caption{ Current frameworks rely on complex agent orchestration with unpredictable execution paths, dramatically increasing API calls and computation time. Their focus on code syntax rather than mathematical accuracy results in costly errors that can propagate through business operations undetected. This excessive coordination overhead makes these systems impractical for time-sensitive business applications. Compared to traditional methods, ORMind employs a streamlined end-to-end workflow with counterfactual reasoning, significantly enhancing solution reliability.}
    \label{fig:ORMind_comparison}
\end{figure*}

Operations research (OR) is critical for business decision-making, helping companies optimize resources, reduce costs, and improve operational efficiency across manufacturing, logistics, and supply chain management. However, previous approaches usually require specialized expertise to translate real-world problems into mathematical optimization problems, hindering their application potential in broader domains. Industry practitioners consistently report that optimization projects face a 30-40\% failure rate due to the disconnect between business requirements and mathematical formulation.

Recent advancements in LLMs have enabled the solving of OR problems. 
Such automation procedures can avoid inconsistent math performance of LLMs~\cite{ahn2024large,imani2023mathprompter,yu2024metamath} and leverage LLMs' ability and knowledge to extract implicit variables and constraints from real-world problems.

However, as Figure~\ref{fig:previous_method} illustrates, existing approaches\cite{xiao2023chain,wang2024unleashing,ahmaditeshniziOptiMUS} to operations research automation face critical deployment challenges. Their complex agent orchestration creates excessive cognitive load through numerous API calls, overwhelming analysts with irrelevant information while significantly increasing costs. These unpredictable expert selection processes reduce solution transparency and create substantial overhead, fundamentally misaligning with human reasoning capabilities.

Inspired by cognitive science and how the brain solves problems, ORMind implements a business-oriented framework based on dual-process theory, combining intuitive analysis with deliberate reasoning. Our specialized modules mirror analyst workflows, from rapid comprehension to deep mathematical thinking. Unlike existing multi-agent frameworks that rely on unpredictable agent selection and complex orchestration, ORMind's innovation lies in its structured, predictable workflow that drastically reduces API calls while maintaining solution quality. ORMind framework is shown in Figure~\ref{fig:our_method}.

We evaluate ORMind on standard benchmark datasets and complex OR problems in industrial scenarios, creating more trustworthy AI systems for business applications. 
Our contributions include:

\begin{itemize}
\item An industry-focused framework that streamlines optimization workflows.
\item A counterfactual reasoning methodology for business-critical constraint validation.
\item A workflow that improves solution trustworthiness and clarity, reducing implementation risks.

\end{itemize}

\section{Related Work}
\label{sec:related}
\noindent\textbf{Operations Research Solving with LLMs.}
Operations research problem solving \cite{nl4opt,ahmaditeshniziOptiMUS,xiao2023chain} contains multiple and diverse applied mathematical problems that require a model to perform complex understanding and reasoning. 
%
A traditional line of approaches \cite{nl4opt} decomposes the OR solving into two separate tasks, first solving the NER task to recognize the optimization problem entities \cite{he2022linearprogrammingwordproblems}, then generating a precise meaning representation of the optimization formulation \cite{gangwar2022tagged}.

Another line of work \cite{tang2024orlm,yang2024benchmarkingllmsoptimizationmodeling} leverages LLMs to synthesize abundant and diverse operations research problems, which later empowers the LLMs with such synthetic data. Such approaches may suffer guaranteed data quality and, at the same time, can be costly.

\noindent\textbf{LLM-based Multi-Agent Workflow}
Recent research has demonstrated the potential of collaborative problem-solving through autonomous cooperation among AI agents \cite{li2023camel,wang2024unleashing,hongmetagpt}. Compared with existing multi-agent collaboration approaches, ORMind’s primary innovation lies in its counterfactual strategy and memory pool coordination mechanism, which aligns more closely with actual business decision-making logic and transparency requirements. This enables the system to exhibit unique advantages in industrial NLP problem scenarios.

\noindent\textbf{LLM-based Reasoning Frameworks.}
Recent advancements in LLMs have introduced various innovative frameworks to enhance their complex reasoning capabilities. 
For example, for solving mathematical problems in such as textbooks and contests \cite{DBLP:journals/corr/abs-2110-14168,DBLP:conf/nips/HendrycksBKABTS21,lightman2023let,DBLP:conf/iclr/ZhengHP22}, current research efforts \cite{goutora,zhu2023solving,yumetamath,hao2024toolkengpt} have explored using LLMs via
employing various structures to enhance reasoning fidelity. 

However, these single-agent reasoning methods demonstrate notable shortcomings when dealing with intricate Operations Research (OR) problems. This is because they struggle to address the combined challenges of implicit constraints and factual hallucination on knowledge-intensive tasks.

\begin{figure*}[t]
    \centering
    \includegraphics[width=\textwidth,keepaspectratio]{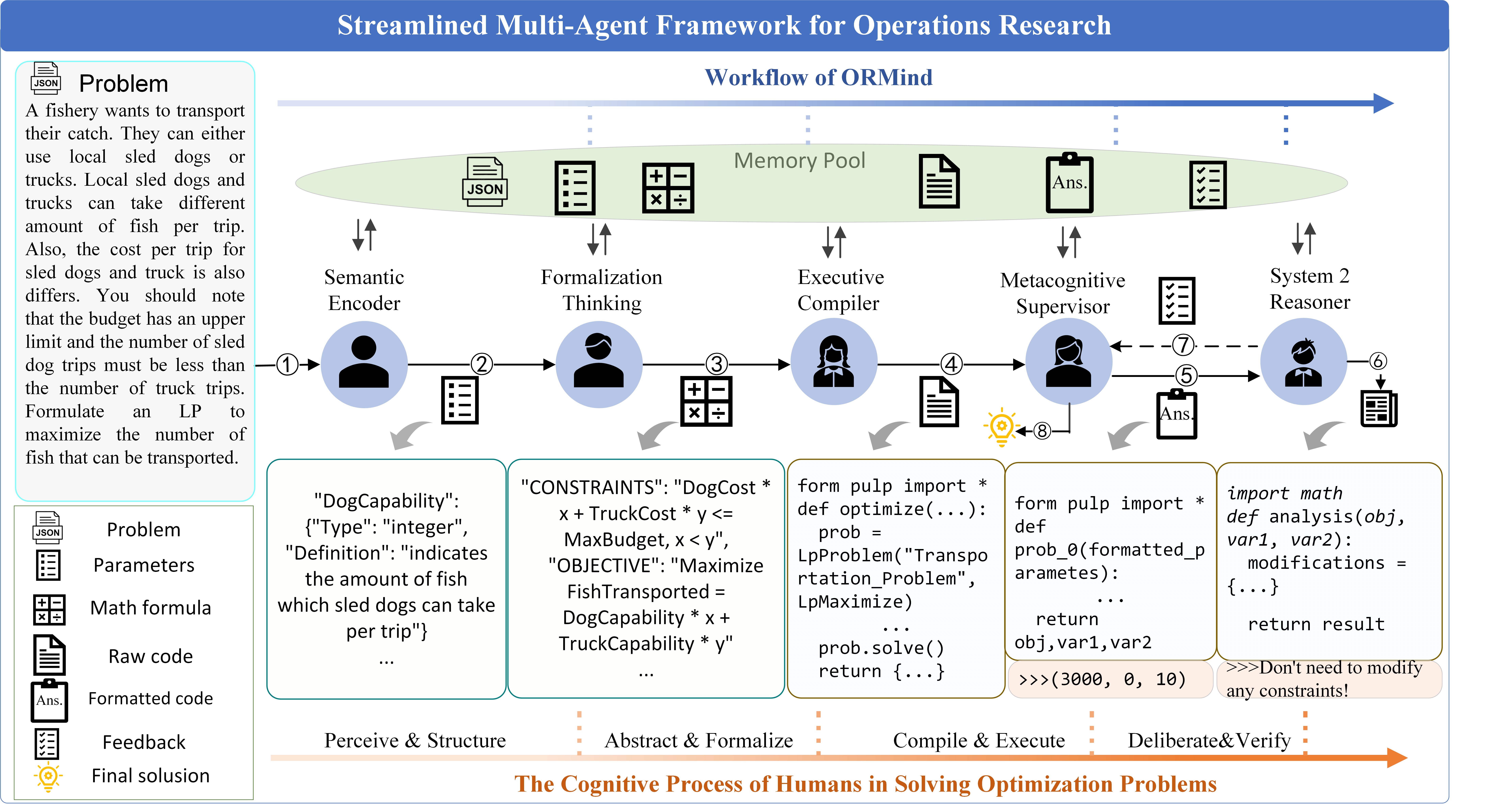} 
    \caption{Our approach is grounded in established cognitive science theories, particularly dual-process framework\cite{kahneman2011thinking} and tripartite model of cognition\cite{stanovich2009distinguishing}. The Semantic Encoder and Formalization Thinking modules correspond to Type 1 (intuitive) processing, while the System 2 Reasoner implements Type 2 (analytical) processing. The Metacognitive Supervisor embodies the reflective mind, monitoring and coordinating between these systems. }
    \label{fig:alg}
\end{figure*}

\section{Methodology}

\subsection{Problem Formulation}
Optimization problems are typically expressed in mathematical terms, consisting of an objective function to be minimized or maximized, subject to a set of constraints. For instance, a Integer Linear Program  can be formulated mathematically as:
\begin{align}
\text{minimize} \quad & \sum_{j=1}^n c_j x_j \\
\text{subject to}  & \sum_{j=1}^n a_{ij} x_j \leq b_i,  i = 1, \ldots, m \\
& l_j \leq x_j \leq u_j, \quad j = 1, \ldots, n \\
& x_j \in \mathbb{Z}, \quad j \in I
\end{align}

\subsection{Architecture Overview}

As illustrated in Figure \ref{fig:alg}, when humans solve optimization problems, their cognitive process aligns with our framework. The brain first performs semantic encoding, rapidly identifying key variables from complex descriptions. It then uses formalization thinking, methodically constructing mathematical relationships between variables and constraints. Next, executive compiler translate these abstract models into actionable solution.

With problem input \( D \) and agent sequence \( \mathbb{A} = \{ A_{\phi_1}, A_{\phi_2}, \ldots, A_{\phi_{N_a}} \} \), where \( N_a \) represents total agents and \( \phi_k \) denotes agent-specific configurations, each component builds upon previous outputs stored in memory pool \( P \).

The transformation operation for agent \( k \) follows:
\[
O_k = A_{\phi_k}(D, P_{k-1})
\]
where \( D \) represents business requirements input and \( P \) contains all previously processed outputs. Each agent's contribution \( O_k \) incrementally enhances the solution repository:
\[
P_k = P_{k-1} \cup \{O_{k}\}
\]

This collaborative memory architecture enables robust business optimization by leveraging specialized expertise while maintaining a comprehensive solution context—critical for enterprise deployments where reliability and solution quality directly impact operational outcomes.

\begin{algorithm}[t]
\caption{Workflow of \ORMind}
\begin{algorithmic}[1]
\Require Pre-processed problems set $\mathbb{D}$=\{$D_1, D_2, \dots, D_{N_T}$\}, maximum number of problems $N_T$, Memory Pool accessible to all modules
\Ensure Optimized solutions $S^*_1, S^*_2, \dots, S^*_{N_T}$
\For{$t = 1$ to $N_T$}
    \State $\Theta_t \gets$ $\mathrm{SemanticEncoder}$($D_t$)
    \State $M_t \gets$ $\mathrm{Formalization}$($D_t$, $\Theta_t$)
    \State $C_t \gets$ $\mathrm{Executive Compiler}$($M_t$)
    \State $F_t \gets$ $\mathrm{Supervisor_{\text{f}}}$($D_t$, $\Theta_t$, $M_t$, $C_t$)
    \State $S_t \gets$ $F_t$ \Comment{Run the code}
    \If{$S_t$ indicates any error}
        \State $R_t \gets$ $\mathrm{Reasoner}$($S_t$,$F_t$)
        \State $F'_t \gets$ $\mathrm{Supervisor}$($D_t$, $\Theta_t$, $M_t$, $C_t$, $R_t$) \Comment{Revise the code based on errors}
        \State $S_t \gets$ $F'_t$ \Comment{Run the code}
    \EndIf
    \State $R_t \gets$ $\mathrm{Reasoner}$($S_t$,$D_t$)
    \If{$R_t$ indicates discrepancies with fact}
        \State $F'_t \gets$ $\mathrm{Supervisor}$($D_t$, $\Theta_t$, $M_t$, $C_t$, $R_t$) 
    \Else
        \State $S^*_t \gets F'_t$ \Comment{Get solution}
    \EndIf
\EndFor
\State \Return $S^*_1, S^*_2, \dots, S^*_{N_T}$
\end{algorithmic}
\end{algorithm}

\subsection{Brief Introduction of Components}

\subsubsection{Semantic Encoder}
The Semantic Encoder transforms unstructured text into structured knowledge representations, reducing the working memory load. It recognizes and categorizes parameters as either scalars or vectors and determines the type of each parameter (e.g., integer, float, boolean, categorical). The output is a parameter set $\Theta = \{\theta_1, \theta_2, ..., \theta_{N_p}\}$, where each $\theta$ represents a parameter with its associated information. This process mirrors the human cognitive ability of selective attention and pattern recognition, where experts rapidly identify and categorize relevant information from complex scenarios.

\subsubsection{Formalization Thinking}
The Formalization Thinking executes deep analytical thinking to construct mathematical models and constraint conditions. The critical steps in this agent involve defining variables, formulating constraints, and constructing the objective function. This component emulates the human brain's abstract reasoning capabilities, where domain experts mentally translate real-world situations into symbolic representations through conceptual abstraction and relationship mapping.

\subsubsection{Executive Compiler}
The Executive Compiler transforms abstract models into executable code snippets $S$, similar to the operationalization process of brain executive functions. 
This transformation reflects the cognitive process of implementation planning, where the human brain converts abstract intentions into concrete action sequences with precise operational details.

\subsubsection{System 2 Reasoner}

System 2 reasoner provides oversight, while deliberate verification employs counterfactual reasoning to test solutions by asking "what if" questions. 
While conventional approaches verify solutions by checking constraints directly, ORMind asks "what constraints need to modify would make this solution optimal?" - essentially learning from hypothetical scenarios to identify potential flaws. This approach mirrors human experts who often validate complex solutions by considering what would need to change for an alternative answer to be correct, enabling more robust error detection than direct verification alone. 
The approach also involves Syntax Error Analysis. In cases where code execution fails due to syntax errors, the specialist pinpoints the problematic line and communicates the probable cause to the Metacognitive Supervisor for swift resolution.

A core innovation in ORMind is the use of counterfactual reasoning for error identification and solution refinement. Assume that the optimization problem can be described by a structural causal model (SCM) with variables $X$, $Y$, and $C$, where:
\begin{align}
Y &= f_Y(X, U), \\
C &= f_C(X, Y, U),
\end{align}
and $U$ denotes latent (exogenous) variables. In our framework, $X$ represents decision variables (e.g., production quantities), $Y$ represents the objective function value (e.g., total cost or profit), and $C$ encapsulates the business constraints.

Inspired by dual-process theories in cognitive science, ORMind divides the reasoning into an intuitive (System~1) phase and a deliberate, analytical (System~2) phase.

For example, given a solution \( S_t = \{obj=150, var_1=30, var_2=20\} \), the System 2 Reasoner might reason:

\begin{align*}
c_1(S_t) &: 2var_1 + 3var_2 \leq 100 \\
c_2(S_t) &: var_1 + var_2 \leq 35
\end{align*}

Using Python tools to assist its reasoning, the agent might determine:
\[
R_t = \begin{cases}
\text{``Modify to: } 2var_1 + 3var_2 \leq 130 \text{''} & \text{for } c_1 \\
\text{``Modify to: } var_1 + var_2 \leq 50 \text{''} & \text{for } c_2
\end{cases}
\]

This approach allows the agent to think through which conditions should be altered to make the given result valid, mimicking the cognitive process of a human expert.

\subsubsection{Metacognitive Supervisor}
The Metacognitive Supervisor mirrors human metacognition—enabling self-awareness of solution quality, strategic oversight, and adaptive decision-making when errors are detected.
It monitors the entire solution generation process, making high-level decision adjustments:
\[
F_t = \mathrm{Supervisor_{\text{forward}}}(D_t, \Theta_t, M_t, C_t)
\]

When constraint violations are detected in production scenarios:
\[
F'_{t} = \mathrm{Supervisor_{\text{backward}}}(S_t, R_t)
\]
where \(R_t\) contains business-critical constraint failure details. The Supervisor uses this intelligence to prioritize adjustments for maximum operational impact.

Once all business constraints are satisfied:
\[
S^*_t = \mathrm{Run}(F'_{t})
\]
This production-ready state \(S^*_{t}\) represents a deployment-vetted solution meeting all business requirements and optimization targets.

\section{Enterprise Application}

Lenovo is piloting this innovative approach within its AI Assistant system. The assistant leverages customer computing requirements and budget constraints to formulate mathematical models that optimize the performance-to-cost ratio. Beyond product configuration, Lenovo's AI Assistant extends this optimization capability throughout the customer journey: it streamlines pre-sale product recommendations to shorten decision cycles, automatically applies maximum discounts during purchases to optimize the ordering process, and efficiently handles post-sale services.

At the same time, ORMind is undergoing internal evaluation to enhance product configurations across 292 product categories comprising more than 8,000 potential SKUs (with approximately 2,000+ active SKUs available for recommendation due to business rules requiring in-stock and direct sales items). During testing, the system handled an average of 3,000+ customer inquiries per day, maintaining configuration time below 6 seconds and achieving task completion rates exceeding 80\%. Internal assessment tracked additional metrics: intent recognition accuracy reached 85\%+, recommendation adoption rate (CTR) was 18\%+, and average customer satisfaction score was 4.2 out of 5. Business analysts found the system's transparent reasoning aligned with their own, enabling quick validation and intervention.

\section{Experiments}

\begin{table*}[htb]
\centering
\footnotesize
\renewcommand{\arraystretch}{1}

\begin{tabular}{l
|ccc|ccc
}
\hline
\small
\multirow{2}{*}{\textbf{Method}} & \multicolumn{3}{c|}{\textbf{NL4Opt}} & \multicolumn{3}{c}{\textbf{ComplexOR}} \\
  & \textbf{SR↑} & \textbf{MFFR↓} & \textbf{IEFR↓} & \textbf{SR↑} & \textbf{MFFR↓} & \textbf{IEFR↓} \\
\hline
tag-BART \cite{gangwar2022tagged}   & 47.9\% & -   & -    & 0\%    & -     & -    \\
\hline
OptiMUS \cite{ahmaditeshniziOptiMUS}    & 28.6\%& 4.0\% & 11.9\% & 9.5\% & 7.9\% & 15.0\%\\
Chain-of-Thought \cite{wei2022chain}      & 45.8\%& 20.5\% & 9.4\%  & 0.5\% & 35.3\% & 8.6\%\\

Progressive Hint \cite{zheng2023progressive}      & 42.1\%& 19.4\% & 10.3\% & 2.2\% & 35.1\% & 13.5\%\\
Tree-of-Thought \cite{yao2024tree}      & 47.3\%& 17.4\% & 9.7\%  & 4.9\% & 31.4\% & 7.6\%\\
Graph-of-Thought \cite{besta2024graph}      & 48.0\%& 16.9\% & 9.1\%  & 4.3\% & 32.4\% & 8.1\%\\
ReAct \cite{yao2023react}                 & 48.5\%& 15.5\% & 11.2\% & 14.6\% & 31.9\% & 10.8\%\\
Reflexion \cite{shinn2023reflexion}            & 50.7\%& 7.3\% & 9.0\%  & 13.5\% & 12.9\% & 10.1\%\\
Solo Performance \cite{wang2024unleashing}     & 46.8\%& 17.9\% & 13.6\% & 7.0\% & 46.5\% & 13.5\%\\

Chain-of-Experts \cite{xiao2023chain}      & 58.9\%& 3.8\% & 7.7\%  & 25.9\% & 7.6\% & 6.4\%\\
\hline
\rowcolor{gray!25}
\textbf{\ORMind} & 68.8\% & 0.4\% & 2.0\% & 40.5\%  & 5.4\% & 21.6\% \\
\hline
\end{tabular}
\caption{Comparison with baselines on Nl4Opt and ComplexOR.}
\label{table:comparison}
\end{table*}

\subsection{Datasets}

To compare our method, we utilized two datasets:

1. \textbf{NL4Opt}: This dataset, collected from the NL4Opt competition\footnote{\url{https://nl4opt.github.io/}} at NeurIPS 2022, contains 1101 elementary-level linear programming (LP) problems. It is divided into 713 training samples, 99 validation samples, and 289 test samples. 

2. \textbf{ComplexOR}: This dataset contains 37 actual industrial optimization problems with the complex constraints and business requirements that characterize real-world applications. Each problem mirrors complex decision-making challenges under various business conditions.

\subsection{Experiment Setup and Metrics}

\begin{table*}[htb]
\centering

\renewcommand{\arraystretch}{1}
\small

\begin{tabular}{l|ccc|ccc}
\hline
\multirow{2}{*}{\textbf{Method}}  & \multicolumn{3}{c|}{\textbf{NL4Opt}} & \multicolumn{3}{c}{\textbf{ComplexOR}} \\
                          & \textbf{SR↑} & \textbf{MFFR↓} & \textbf{IEFR↓} & \textbf{SR↑} & \textbf{MFFR↓} & \textbf{IEFR↓} \\ \hline
\ORMind (Full)  & 68.8\% & 0.4\% & 2.0\% & 40.5\%  & 5.4\% & 21.6\% \\
\hline
w/ Conductor       & 63.2\%     & 0.4 \%     & 1.4\%   & 40.5 \%   & 2.7\%  & 16.2\%            \\
w/ Terminology Interpreter          & 64.9\%           &   0.4\%        & 2.4\%            & 29.7\%             & 5.4\%            & 29.7\%           \\
w/ Code Reviewer             & 33.0\%           & 0.4\%             &  6.6\%          & 32.4\%    
& 0.0\%            & 35.1\%           \\
\hline
w/o Semantic Encoder & 58.0\%  & 1.0\%  & 6.9\%  & 32.4\% & 5.4\% & 24.3\%  \\
w/o Formalization Thinking & 65.6\% & 1.4\% & 7.2\% & 35.1\% & 2.7\% & 32.4\%  \\
w/o Counterfactual Analysis  & 59.4\% & 2.8\% & 11.1\% & 32.4\% & 10.8\% & 24.3\% \\ 
w/o Syntax Error Analysis  & 62.2\% & 1.0\% & 8.3\% & 35.1\% & 5.4\% & 29.7\% \\ 
w/o All modules        & 42.4\%& 18.1\% & 13.2\% & 0.5\% & 36.8\% & 8.6\%\\
\hline
\end{tabular}
\caption{Ablation Study of \ORMind.}
\label{tab:table2}
\end{table*}

We used GPT-3.5-turbo \cite{gpt35} as our default large language model, with a temperature of 0.  Our experimental framework is built upon LangChain\footnote{\url{https://www.langchain.com/}}, an open-source library designed to facilitate the development of applications powered by language models. 
We extend the implementation of \ORMind to other backbones, including GPT-4o-mini and GPT-4 \cite{DBLP:journals/corr/abs-2303-08774}. 

Our evaluation employs metrics that assess both the correctness and executability of solutions against practical requirements:

\textbf{Success Rate (SR):} The success rate in solving problems.

\textbf{Model Formulation Failure Rate (MFFR):} The percentage of optimization problems where the system fails to formulate a valid mathematical model due to constraint interpretation errors.

\textbf{Implementation Execution Failure Rate (IEFR):} The percentage of optimization models that fail during solver execution due to technical incompatibilities or resource limitations.

\begin{figure}[htb]

\centering
\includegraphics[width=0.8\linewidth]{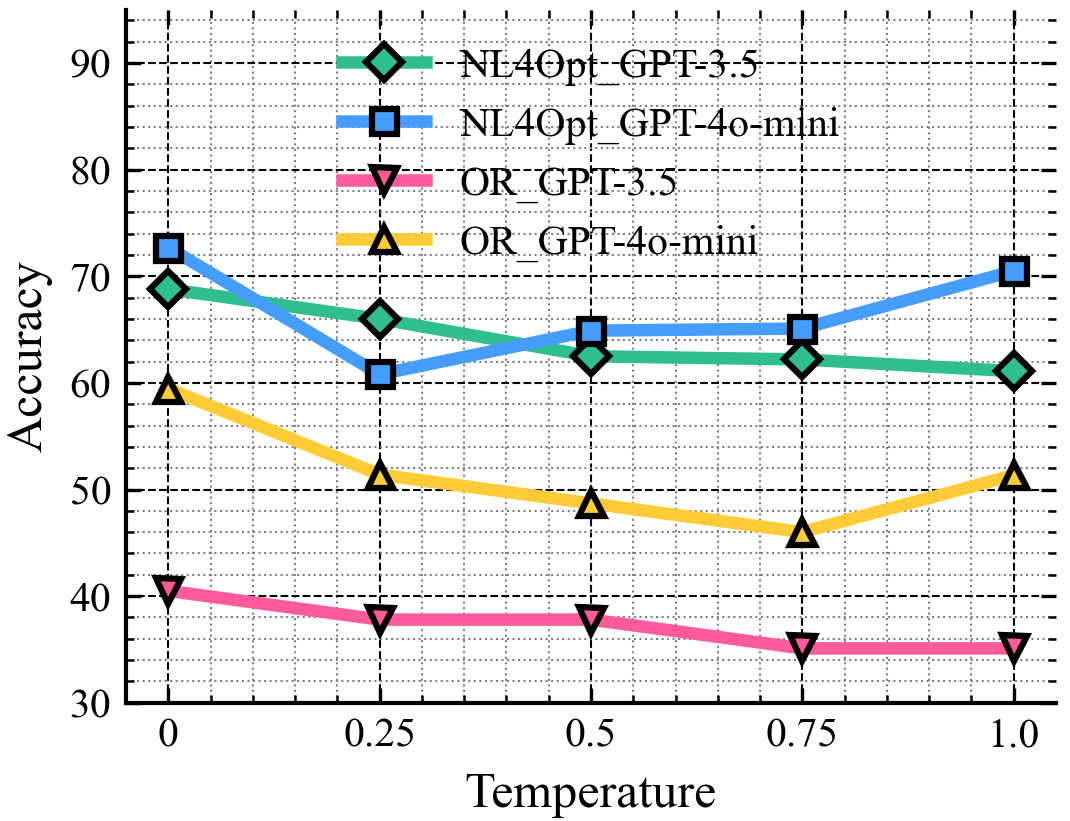}
\caption{Temperature analysis on NL4Opt and ComplexOR}
\label{fig:img2}

\end{figure}

\subsection{Baseline Comparison}

In contrast, \ORMind's more structured, human-inspired workflow provides a clearer and more effective problem-solving strategy, highlighting its advantages in tackling complex operational research challenges.
We benchmark against traditional optimization solutions, including Tag-BART \cite{gangwar2022tagged}, and standard LLM frameworks: Chain-of-Thought \cite{wei2022chain}, Progressive Hint \cite{zheng2023progressive}, Tree-of-Thought \cite{yao2024tree}, Graph-of-Thought \cite{besta2024graph}, ReAct \cite{yao2023react}, Reflexion \cite{shinn2023reflexion}, Solo Performance Prompting \cite{wang2024unleashing}, CoE \cite{xiao2023chain} and OptiMUS \cite{ahmaditeshniziOptiMUS}.

\subsection{Performance Evaluation}

Our evaluation reveals critical limitations in existing approaches. Tag-BART \cite{gangwar2022tagged} completely failed on ComplexOR's complex scenarios, while Reflexion \cite{shinn2023reflexion} showed moderate error-handling capabilities. However, when tackling the more intricate ComplexOR problems, ReAct's performance \cite{yao2023react} slightly surpassed Reflexion, likely due to its advantage in accessing external knowledge bases, underscoring the importance of external data in handling complex scenarios. The results for OptiMUS are cited from their original paper. They suffer significant performance degradation when tested on GPT-3.5 due to counterintuitive workflow structures that deviate from established problem-solving methodologies\cite{ahmaditeshniziOptiMUS}. In practice, we found that the sequence in which agents are invoked in these frameworks often appeared counterintuitive and failed to reflect the natural problem-solving process of human experts. 

The performance disparity between NL4Opt and ComplexOR datasets highlights a key finding: ORMind excels at accurately formulating mathematical models (achieving near-zero MFFR on NL4Opt), while implementation challenges emerge in more complex industrial scenarios (higher IEFR on ComplexOR). This pattern suggests that future improvements should focus on enhancing the robustness of code generation for complex constraint structures rather than model formulation accuracy.

\subsection{Ablation Study}

\subsubsection{Parameter Sensitivity Analysis}

As shown in Figure \ref{fig:img2}, we evaluated the effect of temperature on GPT-3.5 and GPT-4o-mini models. Lower temperature values led to better performance across both models, suggesting that more deterministic expert responses are beneficial.

\begin{table}[htb]
\centering
\small
\renewcommand{\arraystretch}{1}
\begin{tabular}{l|cc}
\hline
\multirow{2}{*}{\textbf{Method}} & \multicolumn{2}{c}{\textbf{GPT-4}}  \\
               & \textbf{NL4Opt} & \textbf{ComplexOR} \\ \hline
Standard    & 47.3\%    & 4.9\%    \\
Reflexion   & 53.0\%   & 16.8\%      \\
Chain-of-Experts   & 64.2\%   & 31.4\%    \\
OptiMUS   & 78.8\%        & 66.7\%     \\
\hline
\rowcolor{gray!25}
\textbf{\ORMind}  & 79.9\% & 62.2\%    \\ 
\hline
\end{tabular}
\caption{Robustness of \ORMind under Different Large Language Models.}
\label{tab:table3}
\end{table}

\subsubsection{Impact of Various Components.} 

Table \ref{tab:table2} quantifies each component's contribution to \ORMind's performance across industry-relevant datasets. Ablation studies show that removing Semantic Encoder or Formalization Thinking significantly reduces solution quality, highlighting their importance for enterprise problem structuring. The System 2 Reasoner proves essential for production systems, with its partial function removal causing 6-9\% performance degradation.

Adding a Conductor for agent selection increased operational complexity without improving performance, as our streamlined approach proved more cost-efficient. Introducing a Terminology Interpreter decreased performance by 3-5\%, suggesting additional interpretation layers create unnecessary overhead. 
Similarly, Code Reviewer caused hallucinations in large language models, incorrectly modifying appropriately functioning code.

\subsubsection{Method Robustness}

Table \ref{tab:table3} demonstrates \ORMind's reliability with GPT-4 as the foundation model. The consistent performance enhancement across metrics confirms that \ORMind's architecture effectively leverages advanced LLMs, delivering superior optimization solutions for business operations.

\subsubsection{Operational Efficiency}

\begin{table}[htb]
\centering
\small
\renewcommand{\arraystretch}{1}
\begin{tabular}{l|c|c}
\hline
Method & NL4Opt & ComplexOR \\
\hline
CoE & $2003 \pm 456$ & $3288 \pm 780$ \\
OptiMUS & $2838 \pm 822$ & $3241 \pm 1194$ \\
\hline
\textbf{\ORMind} & $2676 \pm 518$ & $3336 \pm 997$  \\
\textbf{w/o  Reasoner} & $1539 \pm 228$ & $2390 \pm 500$  \\
\hline
\end{tabular}
\caption{Comparison of prompt lengths across different datasets for other methods.}
\label{tab:table4}
\end{table}

\ORMind maintains optimal token efficiency across enterprise-scale datasets, reducing computational overhead by streamlining earlier processing stages. Ablation study demonstrates that our system exhibits significant robustness, transparency, and engineering efficiency in industrial scenarios.

\section{Conclusion}

This paper introduces ORMind, a cognitive-inspired end-to-end reasoning framework, which is being piloted within Lenovo's AI Assistant as part of internal evaluations to enhance optimization capabilities for business. Future work will validate the framework on larger enterprise datasets and refine module coordination to build a stronger theoretical foundation and practical benchmarks for industrial decision systems.

\section*{Acknowledgement}
\label{sec:ack}

This work was supported by the Scientific Research Innovation Capability Support Project for Young Faculty No.ZYGXQNJSKYCXNLZCXM-I28.

\section*{Ethics Statement}

In developing and deploying the ORMind framework, we have recognized that addressing ethical challenges is crucial for generating fair, transparent, and sustainable outcomes. One of the primary concerns is data bias. To mitigate this risk, we implement rigorous data cleaning and curation processes. Model robustness is another ethical challenge that we address in ORMind. Given the complexity of the multi-agent framework and the heavy reliance on large language models, we recognize that unexpected inputs or adversarial scenarios may lead to instability. As a risk mitigation measure, we have developed a robust error-detection mechanism to catch anomalies and iteratively correct errors.

\section*{Limitations}
Our model’s performance is highly dependent on the input data quality, and even with our robust data cleaning protocols, there is still a risk that residual biases may affect outcomes. Further work is needed to develop automated workflows that periodically audit and adjust data sources, thus reducing this risk over the long term. In terms of robustness, while our multi-agent iterative process allows for continuous refinement, the inherent brittleness of large language models under adversarial conditions poses a challenge. Future improvements will focus on integrating adversarial testing, uncertainty quantification, and more sophisticated error-correction protocols to enhance overall stability. Moreover, the orchestration of multiple agents demands significant computational and memory resources, which may not be feasible in every deployment scenario. To address this issue, we plan to explore model compression, caching techniques, and scalable infrastructure solutions that can dynamically allocate resources based on the current load.

\bibliography{custom}

\appendix
\onecolumn

\section{NL4Opt Case Study}

In this section, we provide a detailed case study to illustrate the workings of our proposed ORMind framework. We take a specific problem from the \textbf{NL4Opt} dataset and walk through the entire process from problem input to final solution. The case study demonstrates how each agent in the ORMind framework contributes to the final output.

\subsection{Problem Description}

\textbf{A pharmacy has 3000 mg of morphine to make painkillers and sleeping pills. Each painkiller pill requires 10 mg of morphine and three units of digestive medicine. Each sleeping pill requires 6 mg of morphine and five units of digestive medicine. The pharmacy needs to make at least 50 painkiller pills. Since sleeping pills are more popular, at least 70\% of the pills should be sleeping pills. How many of each should the pharmacy make to minimize the total amount of digestive medicine needed?}

\subsection{Workflow Execution}


\begin{tcolorbox}[mypromptbox, title=Semantic Encoder]
\begin{lstlisting}[style=promptStyle]
# Function Signature
output = self.forward_chain.invoke(
    {"problem_example": problem_example,
     "comment_text": comment_pool.get_current_comment_text()}
).content

# Output
{
"painkillers": {"Type": "integer", "Definition": "the number of painkiller pills"},
"sleeping_pills": {"Type": "integer", "Definition": "the number of sleeping pills"}
}
\end{lstlisting}
\end{tcolorbox}

\begin{tcolorbox}[mypromptbox, title=Formalization Thinking]
\begin{lstlisting}[style=promptStyle]
# Function Signature
output = self.forward_chain.invoke(
{"problem_description": problem_description,
"comments_text": comments_text}
).content

# Output
{
    "VARIABLES": "painkillers: integer, sleeping_pills: integer",
    "CONSTRAINTS": "painkillers >= 50, sleeping_pills >= 0.7 * (painkillers + sleeping_pills), 10*painkillers + 6*sleeping_pills <= 3000",
    "OBJECTIVE": "Minimize 3*painkillers + 5*sleeping_pills"
}
\end{lstlisting}
\end{tcolorbox}

\begin{tcolorbox}[mypromptbox, title=Executive Compiler]
\begin{lstlisting}[style=promptStyle]
# Function Signature
output = self.forward_chain.invoke({
"problem_description":problem_description,
"comments_text":comments_text}
).content

# Output
import pulp

def optimize_pharmacy():
    # Create a LP minimization problem
    prob = pulp.LpProblem("Pharmacy Optimization", pulp.LpMinimize)
    
    # Define decision variables
    painkillers = pulp.LpVariable("painkillers", lowBound=50, cat='Integer')
    sleeping_pills = pulp.LpVariable("sleeping_pills", lowBound=0, cat='Integer')
    
    # Add constraints
    problem += 10*painkillers_var + 6*sleeping_pills_var <= 3000
    problem += 3*painkillers_var + 5*sleeping_pills_var >= 0.7*(painkillers_var + sleeping_pills_var)
    
    # Set objective function
    prob += 3*painkillers + 5*sleeping_pills
    
    # Solve the problem
    prob.solve()
    
    # Return the optimal values of painkillers and sleeping pills
    return {
        "painkillers": int(painkillers.varValue),
        "sleeping_pills": int(sleeping_pills.varValue)
    }
\end{lstlisting}
\end{tcolorbox}

\begin{tcolorbox}[mypromptbox, title=Metacognitive Supervisor forward]
\begin{lstlisting}[style=promptStyle]
# Function Signature
answer = self.forward_chain.invoke({
    "comment_text": comment_text,
    "code_example":problem["code_example"],
    "attention":attention
}).content

# Output
from pulp import LpProblem, LpVariable, LpMinimize, lpSum

def prob_123(painkillers, sleeping_pills):
    """
    Args:
        painkillers: an integer, representing the number of painkiller pills
        sleeping_pills: an integer, representing the number of sleeping pills
    Returns:
        amount_of_digestive_medicine: an integer, representing the total amount of digestive medicine needed  
    """
    # Create a LP minimization problem
    problem = LpProblem("Pharmacy Optimization Problem", LpMinimize)
    
    # Define decision variables
    painkillers_var = LpVariable("painkillers", lowBound=50, cat='Integer')
    sleeping_pills_var = LpVariable("sleeping_pills", lowBound=0, cat='Integer')
    
    # Add objective function
    problem += 3*painkillers_var + 5*sleeping_pills_var
    
    # Add constraints
    problem += 10*painkillers_var + 6*sleeping_pills_var <= 3000
    problem += 3*painkillers_var + 5*sleeping_pills_var >= 0.7*(painkillers_var + sleeping_pills_var)
    
    # Solve the problem
    problem.solve()
    
    return (problem.objective.value(), int(painkillers_var.varValue), int(sleeping_pills_var.varValue))

    # Running Result:
    (150.0, 50, 0)
\end{lstlisting}
\end{tcolorbox}

\begin{tcolorbox}[mypromptbox, title=System 2 Reasoner]
\begin{lstlisting}[style=promptStyle]
# Function Signature
answer = self.forward_chain.invoke({
    "problem_description": problem['description'],
    "code_example": code_example,
    "input_content": input_content
}).content

# Output
import math

def counterfactual_solution_analysis(obj, var1, var2):
    """
    Analyze what changes would be necessary for the given solution to be valid and optimal.
    The function variable names must remain obj, var1 and var2. Do not alter these names.
    Args:
        obj: The objective value
        var1: Value of variable 1
        var2: Value of variable 2

    Returns:
        dict: Contains suggested modifications for each constraint and overall assessment
    """
    epsilon = 1e-2
    modifications = {
        "Modification1": {
            "check": lambda: var1 >= 0-epsilon,
            "message": "Adjust constraint to allow var1 to be {:.2f}".format(var1)
        },
        "Modification2": {
            "check": lambda: var2 >= 0-epsilon,
            "message": "Adjust constraint to allow var2 to be {:.2f}".format(var2)
        },
        "Modification3": {
            "check": lambda: 10 * var1 + 6 * var2 <= 3000+epsilon,
            "message": "Modify resource constraint to allow 10*var1 + 6*var2 to be {:.2f}".format(10*var1 + 6*var2)
        },
        "Modification4": {
            "check": lambda: var1 >= 50,
            "message": "Adjust constraint to allow var1 to be at least 50"
        },
        "Modification5": {
            "check": lambda: var2 >= 0.7 * (var1 + var2),
            "message": "Adjust constraint to ensure at least 70% of pills are sleeping pills"
        },
        "Modification6": {
            "check": lambda: math.isclose(var1, round(var1)) and math.isclose(var2, round(var2)),
            "message": "Remove integer constraint on variables"
        },
        "Modification7": {
            "check": lambda: math.isclose(obj, round(obj)),
            "message": "Remove integer constraint on objective"
        }
    }

    results = {}
    all_valid = True

    for name, modification in modifications.items():
        needed = not modification["check"]()
        results[name] = {
            "modification_needed": needed,
            "suggestion": modification["message"] if needed else None
        }
        if needed:
            all_valid = False

    results["solution_valid_without_changes"] = all_valid

    return results

    # Running Result:
    Adjust constraint to ensure at least 70% of pills are sleeping pills
\end{lstlisting}
\end{tcolorbox}

\begin{tcolorbox}[mypromptbox, title=Metacognitive Supervisor backward]
\begin{lstlisting}[style=promptStyle]
# Function Signature
output = self.backward_chain.invoke({
"problem_description": problem["description"],
    "previous_code": self.previous_code,
    "feedback": feedback_pool.get_closet_comment_text(),
    "attention": self.attention
}).content

# Output
from pulp import LpProblem, LpVariable, LpMinimize, lpSum

def prob_123(painkillers, sleeping_pills):
    """
    Args:
        painkillers: an integer, representing the number of painkiller pills
        sleeping_pills: an integer, representing the number of sleeping pills
    Returns:
        amount_of_digestive_medicine: an integer, representing the total amount of digestive medicine needed  
    """
    # Create a LP minimization problem
    problem = LpProblem("Pharmacy Optimization Problem", LpMinimize)
    
    # Define decision variables
    painkillers_var = LpVariable("painkillers", lowBound=50, cat='Integer')
    sleeping_pills_var = LpVariable("sleeping_pills", lowBound=0, cat='Integer')
    
    # Add objective function
    problem += 3*painkillers_var + 5*sleeping_pills_var
    
    # Add constraints
    problem += 10*painkillers_var + 6*sleeping_pills_var <= 3000
    problem += 3*painkillers_var + 5*sleeping_pills_var >= 0.7*(painkillers_var + sleeping_pills_var)
    
    # Adjust constraint to ensure at least 70% of pills are sleeping pills
    problem += sleeping_pills_var >= 0.7*(painkillers_var + sleeping_pills_var)
    
    # Solve the problem
    problem.solve()
    
    return (problem.objective.value(), int(painkillers_var.varValue), int(sleeping_pills_var.varValue))


    # Running Result:
(735.0, 50, 117)
\end{lstlisting}
\end{tcolorbox}

\subsection{Discussion of Results}

In this case study, we explored how each agent in the ORMind framework contributed to solving the optimization problem of minimizing the total amount of digestive medicine needed to produce painkillers and sleeping pills at a pharmacy.

Initially, the Semantic Encoder correctly identified key variables, such as the number of painkillers and sleeping pills, as integers. The Formalization Thinking then successfully formulated the problem by defining the constraints and the objective function. Specifically, the constraints ensured that at least 50 painkiller pills must be produced and that at least 70\% of the pills should be sleeping pills, while the objective was to minimize the use of digestive medicine.

The Programming Expert translated this mathematical model into Python code using the `pulp` library, ensuring the formulated constraints were implemented correctly. Upon initial solution generation, the Metacognitive Supervisor evaluated the code and returned a solution where only 50 painkiller pills were produced, with no sleeping pills, resulting in a minimal amount of digestive medicine used. However, this solution did not satisfy the 70\% requirement for sleeping pills.

The System 2 Reasoner identified this issue through counterfactual analysis and suggested adjusting the constraint to enforce the 70\% sleeping pill requirement. After incorporating this feedback, the Metacognitive Supervisor revised the model, leading to a new solution in which 50 painkiller pills and 117 sleeping pills were produced, minimizing the digestive medicine to 735 units.

This iterative process highlights the strength of the ORMind framework in refining solutions through multiple expert agents, each focusing on specific aspects of the problem. By leveraging the System 2 Reasoner’s counterfactual reasoning, the framework was able to correct an oversight in the initial solution, ensuring compliance with all constraints and optimizing the objective function more effectively. This case study demonstrates the framework’s capability to generate solutions and iteratively improve them, thereby achieving a robust and optimal outcome.

\section{ComplexOR Case Study}

In this section, we provide a detailed case study to illustrate the workings of our proposed ORMind framework. We take a specific problem from the \textbf{ComplexOR} dataset and walk through the entire process from problem input to the final solution. The case study demonstrates how each agent in the ORMind framework contributes to the final output.

\subsection{Problem Description}

\textbf{Consider a production problem. Given a set of products param\{ProductNum\}. Each product p is produced at a rate of param\{ProductionRate\_p\} tons per hour. There are param\{AvailableHours\} hours available in a week. The profit per ton for each product p is param\{ProfitPerTon\_p\}. There is a lower limit param\{MinimumSale\_p\} and an upper limit param\{MaximumSale\_p\} on the tons of each product p sold in a week. The problem aims to maximize the total profit from selling all products. It is constrained that the total of hours used by all products may not exceed the hours available. How do we decide the tons of each product p to be produced?}

\subsection{Workflow Execution}

\begin{tcolorbox}[mypromptbox, title=Semantic Encoder]
\begin{lstlisting}[style=promptStyle]
output = self.forward_chain.invoke(
    {"problem_example": problem_example,
     "comment_text": comment_pool.get_current_comment_text()}
).content

# Output
{
    "ProductNum": {"Type": "Integer", "Definition": "The number of products"},
    "ProductionRate": {"Type": "Array of floats, shape: ProductNum", "Definition": "The production rate of each product in tons per hour"},
    "ProfitPerTon": {"Type": "Array of floats, shape: ProductNum", "Definition": "The profit per ton for each product"},
    "MinimumSale": {"Type": "Array of floats, shape: ProductNum", "Definition": "The lower limit on the tons of each product sold in a week"},
    "MaximumSale": {"Type": "Array of floats, shape: ProductNum", "Definition": "The upper limit on the tons of each product sold in a week"},
    "AvailableHours": {"Type": "Integer", "Definition": "The total available hours in a week for production"}
}

\end{lstlisting}
\end{tcolorbox}

\begin{tcolorbox}[mypromptbox, title=Formalization Thinking]
\begin{lstlisting}[style=promptStyle]
output = self.forward_chain.invoke(
{"problem_description": problem_description,
"comments_text": comments_text}
).content

# Output
{
    "VARIABLES": "x_p: Production quantity in tons for product p, where p ranges from 1 to ProductNum",
    "CONSTRAINTS": "1. x_p >= 0 for all p (Non-negativity constraint)\n2. Sum(p=1 to ProductNum) (x_p / ProductionRate_p) <= AvailableHours (Total production time constraint)\n3. MinimumSale_p <= x_p <= MaximumSale_p for all p (Sales constraints)",
    "OBJECTIVE": "Maximize Sum(p=1 to ProductNum) (ProfitPerTon_p * x_p) (Total profit)"
}
\end{lstlisting}
\end{tcolorbox}

\begin{tcolorbox}[mypromptbox, title=Executive Compiler]
\begin{lstlisting}[style=promptStyle]
    \begin{lstlisting}
output = self.forward_chain.invoke({
"problem_description":problem_description,
"comments_text":comments_text}
).content

# Output
import pulp

def solve_production_problem(ProductNum, ProductionRate, ProfitPerTon, MinimumSale, MaximumSale, AvailableHours):
    # Create a linear programming problem to maximize profit
    problem = pulp.LpProblem("Maximize_Profit", pulp.LpMaximize)
    
    # Define decision variables
    x = pulp.LpVariable.dicts("Production_Quantity", range(ProductNum), lowBound=0)
    
    # Objective Function: Maximize total profit
    profit_contributions = [ProfitPerTon[i] * x[i] for i in range(ProductNum)]
    problem += pulp.lpSum(profit_contributions), "Total_Profit"
    
    # Constraint: Total production time must not exceed available hours
    time_contributions = [(x[i] / ProductionRate[i]) for i in range(ProductNum)]
    problem += pulp.lpSum(time_contributions) <= AvailableHours, "Total_Production_Time"
    
    # Constraints: Sales constraints for each product
    for i in range(ProductNum):
        problem += x[i] >= MinimumSale[i], f"Min_Sale_{i}"
        problem += x[i] <= MaximumSale[i], f"Max_Sale_{i}"
    
    # Solve the problem
    problem.solve()
    
    # Check if a valid solution has been found
    if pulp.LpStatus[problem.status] == 'Optimal':
        # Extract the solution
        production_quantities = {f'Product_{i+1}': x[i].varValue for i in range(ProductNum)}
        total_profit = pulp.value(problem.objective)
        return production_quantities, total_profit
    else:
        return None, None
 \end{lstlisting}
\end{tcolorbox}

\begin{tcolorbox}[mypromptbox, title=Metacognitive Supervisor(forward)]
\begin{lstlisting}[style=promptStyle]
answer = self.forward_chain.invoke({
    "comment_text": comment_text,
    "code_example":problem["code_example"],
    "attention":attention
}).content

# Output
from pulp import *
import numpy as np

def solve(data):
    ProductNum = data["ProductNum"]
    ProductionRate = np.array(data["ProductionRate"])
    ProfitPerTon = np.array(data["ProfitPerTon"])
    MinimumSale = np.array(data["MinimumSale"])
    MaximumSale = np.array(data["MaximumSale"])
    AvailableHours = data["AvailableHours"]

    prob = LpProblem("Steel3_Problem", LpMaximize)
    tons = LpVariable.dicts("Tons", range(ProductNum), lowBound=0, cat='Continuous')

    # Objective Function: Maximize total profit
    prob += lpSum([ProfitPerTon[i] * tons[i] for i in range(ProductNum)])

    # Constraint: Total production time must not exceed available hours
    prob += lpSum([tons[i] / ProductionRate[i] for i in range(ProductNum)]) <= AvailableHours

    # Constraints: Sales constraints for each product
    for i in range(ProductNum):
        prob += tons[i] >= MinimumSale[i]
        prob += tons[i] <= MaximumSale[i]

    prob.solve()

    if prob.status == LpStatusOptimal:
        # Extract the optimized variables as values
        optimized_vars = {
            "tons": {i: tons[i].varValue for i in range(ProductNum)}
        }
        return {
            "status": "Optimal",
            "objective_value": value(prob.objective),
            "optimized_vars": optimized_vars
        }
    else:
        return {"status": LpStatus[prob.status]}

    # Running Result:
TypeError: unsupported operand type(s) for /: 'LpVariable' and 'int'
\end{lstlisting}
\end{tcolorbox}

\begin{tcolorbox}[mypromptbox, title=System 2 Reasoner]
\begin{lstlisting}[style=promptStyle]
answer = self.forward_chain.invoke({
    "problem_description": problem['description'],
    "code_example": code_example,
    "input_content": input_content
}).content

# Output
1. The error is caused because the PuLP library does not support division of an `LpVariable` by an integer directly in the constraint formulation.
2. The specific problematic code section is: `prob += lpSum([tons[i] / ProductionRate[i] for i in range(ProductNum)]) <= AvailableHours`.
\end{lstlisting}
\end{tcolorbox}

\begin{tcolorbox}[mypromptbox, title=Metacognitive Supervisor(backward)]
\begin{lstlisting}[style=promptStyle]
output = self.backward_chain.invoke({
"problem_description": problem["description"],
    "previous_code": self.previous_code,
    "feedback": feedback_pool.get_closet_comment_text(),
    "attention": self.attention
}).content

# Output
from pulp import *

def solve(data):
    ProductNum = data["ProductNum"]
    ProductionRate = data["ProductionRate"]
    ProfitPerTon = data["ProfitPerTon"]
    MinimumSale = data["MinimumSale"]
    MaximumSale = data["MaximumSale"]
    AvailableHours = data["AvailableHours"]

    prob = LpProblem("Steel3_Problem", LpMaximize)
    tons = LpVariable.dicts("Tons", range(ProductNum), lowBound=0, cat='Continuous')

    # Objective Function: Maximize total profit
    prob += lpSum([ProfitPerTon[i] * tons[i] for i in range(ProductNum)])

    # Constraint: Total production time must not exceed available hours
    # Correcting the division issue by multiplying with ProductionRate[i] instead of dividing
    prob += lpSum([tons[i] * (1 / ProductionRate[i]) for i in range(ProductNum)]) <= AvailableHours

    # Constraints: Sales constraints for each product
    for i in range(ProductNum):
        prob += tons[i] >= MinimumSale[i]
        prob += tons[i] <= MaximumSale[i]

    prob.solve()

    if prob.status == LpStatusOptimal:
        # Extract the optimized variables as values
        optimized_vars = {
            "tons": {i: tons[i].varValue for i in range(ProductNum)}
        }
        return {
            "status": "Optimal",
            "objective_value": value(prob.objective),
            "optimized_vars": optimized_vars
        }
    else:
        return {"status": LpStatus[prob.status]}


    # Running Result:
{'status': 'Optimal', 'objective_value': 194828.5706, 'optimized_vars': {'tons': {0: 6000.0, 1: 500.0, 2: 1028.5714}}}
\end{lstlisting}
\end{tcolorbox}

\subsection{Discussion of Results}

The process began with the Semantic Encoder correctly identifying the key variables, such as the number of products, production rates, profits per ton, and sales constraints. These parameters were crucial in formulating the problem accurately.

Next, the Formalization Thinking constructed the mathematical model by defining the decision variables and the constraints. The objective function was set to maximize the total profit. At the same time, the constraints ensured that the total production time did not exceed the available hours and that the production quantities stayed within the specified sales limits.

The Programming Expert then translated this model into Python code using the pulp library. This initial code successfully captured the essence of the problem but encountered a technical issue: the division of LpVariable objects by integers within the constraints, which the pulp library does not directly support.

The System 2 Reasoner identified this issue and provided specific feedback, pinpointing the problematic code and the nature of the error. This feedback was crucial in guiding the Metacognitive Supervisor's subsequent code revision.

The Metacognitive Supervisor corrected the division issue by multiplying instead of dividing the variables within the constraint formulation. This adjustment ensured that the constraints were correctly implemented and allowed the model to be solved without errors.

Finally, the revised model was solved, yielding an optimal solution where the production quantities and total profit were maximized while adhering to all constraints. The solution indicated optimal production quantities for each product and a corresponding total profit, demonstrating the effectiveness of the ORMind framework.

\section{Prompt Templates for Agents}

Below, we list the prompt templates used for each agent in the ORMind framework. These templates are crucial for guiding the LLMs in performing their respective tasks.

\begin{tcolorbox}[mypromptbox, title=Semantic Encoder]
\begin{lstlisting}[style=promptStyle]

#  Prompt Template:
Please review the following example and extract the parameters along with their concise definitions:
{problem_example}
The comment from your colleague is:
{comment_text}
Your output should be in JSON format as follows:
{{
    "Parameter1": {{"Type": "The parameter's data type or shape", "Definition": "A brief definition of the parameter"}},
    "Parameter2": {{"Type": "The parameter's data type or shape", "Definition": "A brief definition of the parameter"}},
    ...
}}
Provide only the requested JSON output without any additional information.
\end{lstlisting}
\end{tcolorbox}

\begin{tcolorbox}[mypromptbox, title=Formalization Thinking]
\begin{lstlisting}[style=promptStyle]

#  Prompt Template:
Now the origin problem is as follows:
{problem_description}
You can use the parameters information from your colleague:
{comments_text}
The order of given parameters is random. You should clarify the meaning of each parameter to choose proper parameter to construct constraint.
Give your Mathematical model of this problem.
Your output format should be a JSON like this:
{{
    "VARIABLES": "A concise description about variables and its shape or type",
    "CONSTRAINTS": "A mathematical Formula about constraints",
    "OBJECTIVE": "A mathematical Formula about objective"
}}
Don't give any other information.
\end{lstlisting}
\end{tcolorbox}

\begin{tcolorbox}[mypromptbox, title=Executive Compiler]
\begin{lstlisting}[style=promptStyle]

#  Prompt Template:
You are presented with a specific problem and tasked with developing an efficient Python program to solve it.
The original problem is as follows:
{problem_description}
Your colleague has constructed a mathematical model for reference:
{comments_text}
Please note that this model may contain errors and is used as a reference. 
You can analyze the problem step by step and provide your own code.
Requirements:
1. Use the PuLP library for implementation.
2. Provide a function that solves the problem.
3. Do not include code usage examples or specific variable values.
4. Focus on creating a general, reusable solution.
\end{lstlisting}
\end{tcolorbox}

\begin{tcolorbox}[mypromptbox, title=System 2 Reasoner]
\begin{lstlisting}[style=promptStyle]

#  Prompt Template:
Analyze the following optimization problem:
{problem_description}

Task: Write a Python function that identifies which specific constraints or conditions in the given problem are not satisfied. This condition will need modification to achieve a valid and optimal solution.

Function specifications:
- Input arguments and their types: {input_content}
- Adhere to the given data types.
- Reference this code structure: {code_example}
- Import the necessary libraries.

Notes:
The code example is only for reference in terms of format and structure. Generate code specifically for the given problem, not based on any examples. 
All specific constraints should be determined based on the problem description provided. 
Make sure to include checks for all constraints mentioned in the problem description. Don't give any Example usages.
\end{lstlisting}
\end{tcolorbox}

\begin{tcolorbox}[mypromptbox, title=Metacognitive Supervisor(backward)]
\begin{lstlisting}[style=promptStyle]

#  Prompt Template:
    FORWARD_TASK: Your colleague Executive Compiler has given his answer:
{comment_text}
This answer has not been formatted. You need to format the code as the example.
The final code must has the same input args and function name as the code example:
{code_example}
You also need to return the optimized variables. 
Important: Your final code should strictly use same input args, function name and return style of the code example exactly.
{attention}
Don't forget to import the library. Don't give any example usage.

    BACKWARD_TASK: In your previous answer may have errors, you receive feedback about the error.
The feedback is generated by counterfactual reasoning, 
which means that the feedback does not represent actual changes that need to be made to the problem conditions.
the feedback highlights where your code may have misinterpreted the original conditions.
{feedback}

For example, If you receive a message like 'Remove integer constraint on variables',
it means that your previous answer is correct only when the integer constraint is removed. 
This strongly suggests that your earlier solution likely overlooked the integer constraint. 
You need to add the constraint.
If you receive a message like 'Modify resource constraint to allow...',
it means that your previous answer is correct only when this constraint is modified. 
This strongly suggests that your earlier solution likely has error in this constraint.
You need to doublecheck your previous code corresponding to the feedback and fix the error.

Carefully review the feedback and give the final code as the format of your previous code.
{attention}

The original problem description remains unchanged:
{problem_description}

Your previous code for analyzing the solution was:
{previous_code}

Your task is to carefully review the original problem description and the counterfactual feedback.

Remember:
Provide your corrected code in the same format as your previous code.
Do not give any example or explanation.
If the feedback is not existed in the description, you may directly use the original code.
Use "from PuLP import *" to import the library as the example.
\end{lstlisting}
\end{tcolorbox}

\begin{tcolorbox}[mypromptbox, title=Conductor]
\begin{lstlisting}[style=promptStyle]

#  Prompt Template:
Now, you are presented with an operational optimization-related problem: 
{problem_description}
In this multi-expert system, there are many agent_team, each of whom is responsible for solving part of the problem.
Your task is to CHOOSE THE NEXT EXPERT TO CONSULT.
The names of the agent_team and their capabilities are listed below:
{experts_info} 
Experts that have already been commented include: 
{commented_experts}
Please select an expert to consult from the remaining expert names {remaining_experts}.
Please note that the maximum number of asked agent_team is {max_collaborate_nums}, and you can ask {remaining_collaborate_nums} more times.
You should output the name of expert directly. The next expert is:'''
\end{lstlisting}
\end{tcolorbox}

\begin{tcolorbox}[mypromptbox, title=Terminology Interpreter]
\begin{lstlisting}[style=promptStyle]

#  Prompt Template:
As a domain knowledge terminology interpreter, your role is to provide additional information and insights related to the problem domain. 
Here are some relevant background knowledge about this problem: {knowledge}. 

You can contribute by sharing your expertise, explaining relevant concepts, and offering suggestions to improve the problem understanding and formulation. 
Please provide your input based on the given problem description: 
{problem_description}

Your output format should be a JSON like this (choose at most 3 hardest terminology. Please provide your output, ensuring there is no additional text or formatting markers like ```json. The output should be in plain JSON format, directly parsable by json.loads(output).):
[
  {{
    "terminology": "...",
    "interpretation": "..."
  }}
]
\end{lstlisting}
\end{tcolorbox}

\begin{tcolorbox}[mypromptbox, title=Code Reviewer]
\begin{lstlisting}[style=promptStyle]

#  Prompt Template:
As a Code Reviewer, your responsibility is to conduct thorough reviews of implemented code related to optimization problems. 
You will identify possible errors, inefficiencies, or areas for improvement in the code, ensuring that it adheres to best practices and delivers optimal results. Now, here is the problem: 
{problem_description}. 

You are supposed to refer to the codes given by your colleagues from other aspects: {comments_text}
\end{lstlisting}
\end{tcolorbox}

\section{Code Example}

The following are code examples used by the ORMind framework for the Counterfactual Analysis.

\begin{tcolorbox}[mypromptbox, title=NL4Opt Code Example for Counterfactual Analysis]
\begin{lstlisting}[style=promptStyle]
import math

def counterfactual_solution_analysis(obj, var1, var2):
    """
    Analyze what changes would be necessary for the given solution to be valid and optimal.
    The function variable names must remain obj, var1 and var2. Do not alter these names.
    Args:
        obj: The objective value
        var1: Value of variable 1
        var2: Value of variable 2

    Returns:
        dict: Contains suggested modifications for each constraint and overall assessment
    """
    epsilon = 1e-2
    modifications = {
        "Modification1": {
            "check": lambda: var1 >= 0-epsilon,
            "message": "Adjust constraint to allow var1 to be {:.2f}".format(var1)
        },
        "Modification2": {
            "check": lambda: var2 >= 0-epsilon,
            "message": "Adjust constraint to allow var2 to be {:.2f}".format(var2)
        },
        "Modification3": {
            "check": lambda: 2 * var1 + 3 * var2 <= 100+epsilon,
            "message": "Modify resource constraint to allow 2*var1 + 3*var2 to be {:.2f}".format(2*var1 + 3*var2)
        },
        "Modification4": {
            "check": lambda: var1 + var2 <= 35+epsilon,
            "message": "Adjust daily production limit to allow var1 + var2 to be {:.2f}".format(var1 + var2)
        },
        "Modification5": {
            "check": lambda: math.isclose(var1, round(var1)) and math.isclose(var2, round(var2)),
            "message": "Remove integer constraint on variables"
        },
        "Modification6": {
            "check": lambda: math.isclose(obj, round(obj)),
            "message": "Remove integer constraint on objective"
        }
    }

    results = {}
    all_valid = True

    for name, modification in modifications.items():
        needed = not modification["check"]()
        results[name] = {
            "modification_needed": needed,
            "suggestion": modification["message"] if needed else None
        }
        if needed:
            all_valid = False

    results["solution_valid_without_changes"] = all_valid

    return results
\end{lstlisting}
\end{tcolorbox}

\begin{tcolorbox}[mypromptbox, title=ComplexOR Code Example for Counterfactual Analysis]
\begin{lstlisting}[style=promptStyle]
import numpy as np

def counterfactual_solution_analysis(alloys_used, data):
    """
    Analyze what changes would be necessary for the given solution to be valid and optimal.

    Returns:
        dict: Contains suggested modifications for each constraint and overall assessment
    """
    AlloysOnMarket = data["AlloysOnMarket"]
    RequiredElements = data["RequiredElements"]
    CompositionDataPercentage = np.array(data["CompositionDataPercentage"])
    DesiredBlendPercentage = np.array(data["DesiredBlendPercentage"])
    AlloyPrice = np.array(data["AlloyPrice"])

    alloys_used_array = np.array([alloys_used[a] for a in range(AlloysOnMarket)])

    modifications = {
        "Modification1": {
            "check": lambda: all(alloys_used_array >= 0),
            "message": "Adjust non-negativity constraint to allow negative quantities: {}".format(alloys_used_array)
        },
        "Modification2": {
            "check": lambda: all(np.dot(CompositionDataPercentage, alloys_used_array) >= DesiredBlendPercentage * np.sum(alloys_used_array)),
            "message": "Modify desired blend percentages to: {}".format(np.dot(CompositionDataPercentage, alloys_used_array) / np.sum(alloys_used_array))
        },
        "Modification3": {
            "check": lambda: all(alloys_used_array <= 1),
            "message": "Increase market availability to allow quantities: {}".format(alloys_used_array)
        }
    }

    results = {}
    all_valid = True

    for name, modification in modifications.items():
        needed = not modification["check"]()
        results[name] = {
            "modification_needed": needed,
            "suggestion": modification["message"] if needed else None
        }
        if needed:
            all_valid = False

    results["solution_valid_without_changes"] = all_valid

    return results
\end{lstlisting}
\end{tcolorbox}
\section{Hardware and Software Configurations}



\subsection{Software}

The software environment used in the experiments includes:
- \textbf{Operating System}: Windows11
- \textbf{Python}: 3.10
- \textbf{LangChain}: 0.2.7
- \textbf{LangChain-Community}: 0.2.7
- \textbf{NumPy}: 1.23.5
- \textbf{Tqdm}: 4.62.3
- \textbf{PuLP}: 2.8.0
- \textbf{OpenAI API Key}: Required for accessing OpenAI's models

\section{Data Format Example}

\begin{tcolorbox}[mypromptbox, title= Formatted NL4OPT data in JSON format]
\begin{lstlisting}[style=promptStyle]


"description":A fishery wants to transport their catch. They can either use local sled dogs or trucks. Local sled dogs and trucks can take different amount of fish per trip. Also, the cost per trip for sled dogs and truck is also differs. You should note that the budget has an upper limit and the number of sled dog trips must be less than the number of truck trips. Formulate an LP to maximize the number of fish that can be transported.
[
    {
        "input": {
            "DogCapability": 100,
            "TruckCapability": 300,
            "DogCost": 50,
            "TruckCost": 100,
            "MaxBudget": 1000
        },
        "output": [
            3000
        ]
    }
]

\end{lstlisting}
\end{tcolorbox}

\begin{tcolorbox}[mypromptbox, title=Formatted ComplexOR data in JSON format]
\begin{lstlisting}[style=promptStyle]

{
    "description": "The Aircraft Assignment Problem is a mathematical programming model that aims to assign \\param{TotalAircraft} aircraft to \\param{TotalRoutes} routes in order to minimize the total cost while satisfying availability and demand constraints. The availability for each aircraft i is \\param{Availability_i} and it represents the maximum number of routes that the aircraft can be assigned to. The demand for each route j is \\param{Demand_j} and it denotes the number of aircraft required to fulfill the passenger or cargo needs of the route. The capability of each aircraft i for each route j is given by \\param{Capacity_{i,j}} and it defines whether the aircraft can service the route, considering factors such as range, size, and suitability. Finally, \\param{Cost_{i,j}} represents the cost of assigning aircraft i to route j, which includes operational, fuel, and potential opportunity costs.",
    "parameters": [
        {
            "symbol": "TotalAircraft",
            "definition": "The total number of aircraft available for assignment",
            "shape": []
        },
        {
            "symbol": "TotalRoutes",
            "definition": "The total number of routes that require aircraft assignment",
            "shape": []
        },
        {
            "symbol": "Availability",
            "definition": "The availability of each aircraft, indicating the maximum number of routes it can be assigned to",
            "shape": [
                "TotalAircraft"
            ]
        },
        {
            "symbol": "Demand",
            "definition": "The demand for each route, indicating the number of aircraft required",
            "shape": [
                "TotalRoutes"
            ]
        },
        {
            "symbol": "Capacity",
            "definition": "The capacity matrix defining the suitability and capability of each aircraft for each route",
            "shape": [
                "TotalAircraft",
                "TotalRoutes"
            ]
        },
        {
            "symbol": "Costs",
            "definition": "The cost matrix representing the cost of assigning each aircraft to each route",
            "shape": [
                "TotalAircraft",
                "TotalRoutes"
            ]
        }
    ]
}



[
{
    "TotalAircraft": 5,
    "TotalRoutes": 5,
    "Availability": [10, 19, 25, 15, 0],
    "Demand": [250, 120, 180, 90, 600],
    "Capacity": [
        [16, 15, 28, 23, 81],
        [0, 10, 14, 15, 57],
        [0, 5, 0, 7, 29],
        [9, 11, 22, 17, 55],
        [1, 1, 1, 1, 1]
    ],
    "Costs": [
        [17, 5, 18, 17, 7],
        [15, 20, 9, 5, 18],
        [15, 13, 8, 5, 19],
        [13, 14, 6, 16, 8],
        [13, 14, 14, 10, 7]
    ]
},
    "output": [
        "Infeasible"
        ]
    }
]
\end{lstlisting}
\end{tcolorbox}

\section{Agent-Memory Pool Interaction in ORMind}
\label{sec:memory_pool_interaction}

The Memory Pool in \textit{ORMind} functions as a centralized repository that supports the collaboration and coordination of agents during the reasoning process. It stores and provides access to shared data, ensuring consistency and efficiency in solving complex operations research (OR) problems.

Agents interact with the Memory Pool primarily through retrieval and update. Before performing a task, an agent retrieves relevant information from the Memory Pool, such as the current problem state, previously identified variables and constraints, and intermediate results from earlier reasoning steps. This ensures that all agents operate with access to the most up-to-date context, avoiding redundant computations and inconsistencies.

Once an agent completes a task, it updates the Memory Pool with its results. These updates include newly discovered variables, constraints, other task-specific outputs, and annotations summarizing the reasoning process. Every update is tagged with metadata, such as the agent's identifier and a timestamp, to maintain traceability and facilitate debugging.

The Memory Pool also plays a critical role in the iterative refinement process. As new information becomes available, earlier results can be revisited and improved by subsequent agents, allowing for modular and adaptive problem-solving. This centralized structure ensures that the system's collective progress is reflected in a single shared repository, enabling efficient and coherent reasoning across all agents.

The Memory Pool enhances the ORMind framework's ability to tackle complex OR problems by providing a shared, structured, and continuously updated context. It promotes collaboration, reduces redundancy, and ensures that agents work synchronized and context-awarely.

\section{Comparison with Other Planning with Feedback Methods}

While our methodology adopts a multi-expert framework, it distinguishes itself through two unique features: human problem-solving process and counterfactual reasoning. These features enable a more structured and iterative problem-solving process compared to other approaches.

The table \ref{tab:comparison} highlights the differences between our approach and other methods regarding key functionalities such as multi-agent frameworks, industry-focused processes, external knowledge access, and feedback refinement.


\begin{table*}[htb]
\centering
\small 
\setlength{\tabcolsep}{4pt} 
\begin{tabular}{lcccc}
\hline
\textbf{Method} & \textbf{Multi-agents} & \textbf{Industry-Focused } & \textbf{External Knowledge} & \textbf{Feedback Refinement} \\
\hline
ReAct\cite{yao2023react} & \xmark & \xmark & \checkmark & \xmark \\
Voyager\cite{wangvoyager} & \xmark & \xmark & \checkmark & \checkmark \\
Ghost\cite{zhu2023ghost} & \xmark & \xmark & \checkmark & \checkmark \\
SayPlan\cite{rana2023sayplan} & \xmark & \xmark & \checkmark & \checkmark \\
MetaGPT\cite{hong2024metagpt} & \checkmark & \xmark & \checkmark & \xmark \\
NLSOM\cite{zhuge2023mindstorms} & \checkmark & \xmark & \checkmark & \xmark \\
SSP\cite{wang2024unleashing} & \checkmark & \xmark & \xmark & \xmark \\
ChatEval\cite{chanchateval} & \checkmark & \xmark & \xmark & \xmark \\
ORMind & \checkmark & \checkmark & \xmark & \checkmark \\
\hline
\end{tabular}
\caption{Comparison of ORMind with existing planning and feedback-based methods.}
\label{tab:comparison}
\end{table*}

\section{Long-term Research Value and Future Directions}
\label{sec:longterm}

This work establishes a foundation for advanced reasoning systems in operations research with implications far beyond the current implementation. Below, we analyze the key long-term research values and potential future directions:

\subsection{Counterfactual Reasoning as a Fundamental AI Capability}
\label{subsec:counterfactual}

The counterfactual reasoning approach introduced in ORMind represents a fundamental advancement in how AI systems can validate and refine solutions. By reasoning about what constraints would need to change for a given solution to be valid, our approach begins to bridge the gap between correlation and causation in AI reasoning systems. This opens avenues for more sophisticated causal reasoning frameworks that can identify patterns and underlying causal mechanisms. Beyond operations research, this methodology could fundamentally transform how AI systems approach problem validation and solution refinement across domains ranging from scientific discovery to medical diagnosis. The ability to perform "what-if" analyses on potential solutions provides a form of self-verification that increases solution reliability without requiring explicit programming of all edge cases, a crucial advancement for mission-critical enterprise applications.

\subsection{Cognitive Architectures for Complex Decision Making}
\label{subsec:cognitive}

ORMind's cognitively-inspired framework mirrors human expert reasoning processes and offers a blueprint for next-generation business intelligence systems. The sequential decomposition of complex problems into stages of understanding, formulation, and refinement provides a generalizable architecture that could be applied to various reasoning tasks beyond optimization. This represents a significant shift from current approaches that often rely on monolithic models or rigid predefined workflows. Future research could explore how such cognitive architectures can dynamically adapt their reasoning strategies based on problem characteristics, incorporate domain-specific knowledge while preserving flexible reasoning, and create natural interaction points for human-AI collaboration. The emergence of such cognitively-aligned systems could fundamentally transform how organizations approach complex decision-making, enabling more intuitive, explainable, and effective enterprise AI solutions.

\end{document}